%% file: 0-main-kdd25.tex
\documentclass[sigconf]{acmart}
\input{0.5-usepacks}

\copyrightyear{2025}
\acmYear{2025}
\setcopyright{acmlicensed}\acmConference[KDD '25]{Proceedings of the 31st ACM SIGKDD Conference on Knowledge Discovery and Data Mining V.1}{August 3--7, 2025}{Toronto, ON, Canada}
\acmBooktitle{Proceedings of the 31st ACM SIGKDD Conference on Knowledge Discovery and Data Mining V.1 (KDD '25), August 3--7, 2025, Toronto, ON, Canada}
\acmDOI{10.1145/3690624.3709242}
\acmISBN{979-8-4007-1245-6/25/08}

\begin{document}

\title{Handling Feature Heterogeneity with Learnable Graph Patches}

\author{Yifei Sun}
\affiliation{%
  \institution{Zhejiang University}
  \city{Hangzhou}
  \country{China}
}
\email{yifeisun@zju.edu.cn}

\author{Yang Yang}
\authornote{Corresponding author.}
\affiliation{%
  \institution{Zhejiang University}
  \city{Hangzhou}
  \country{China}
}
\email{yangya@zju.edu.cn}

\author{Xiao Feng}
\affiliation{%
  \institution{Zhejiang University}
  \city{Hangzhou}
  \country{China}
}
\email{functionendless@zju.edu.cn}

\author{Zijun Wang}
\affiliation{%
  \institution{Zhejiang University}
  \city{Hangzhou}
  \country{China}
}
\email{3200103994@zju.edu.cn}

\author{Haoyang Zhong}
\affiliation{%
  \institution{Huazhong University of Science and Technology}
  \city{Wuhan}
  \country{China}
}
\email{haoyangzhong@hust.edu.cn}

\author{Chunping Wang}
\affiliation{%
  \institution{Finvolution Group}
  \city{Shanghai}
  \country{China}
}
\email{wangchunping02@xinye.com}

\author{Lei Chen}
\affiliation{%
  \institution{Finvolution Group}
  \city{Shanghai}
  \country{China}
}
\email{chenlei04@xinye.com}

\renewcommand{\shortauthors}{Yifei Sun et al.}

\input{1-abs}

\ccsdesc[500]{Networks~Network algorithms}

\keywords{Graph neural networks, graph pre-training, feature heterogeneity}

\maketitle

\input{2-intro}
\input{3-related}
\input{4-model}

\input{5-exp}
\input{6-conclusion}

\newpage
\bibliographystyle{ACM-Reference-Format}
\balance
\bibliography{sample-base}

\appendix
\input{7-appendix}

\end{document}

%% file: 0.5-usepacks.tex
\usepackage{xcolor}         
\usepackage{xspace}
\usepackage{graphicx}
\usepackage{float}

\usepackage{booktabs}

\usepackage{paralist}
\usepackage{wrapfig}
\usepackage{amsmath}
\usepackage{threeparttable}
\usepackage{multirow}
\usepackage{algorithm}
\usepackage{diagbox}

\newcommand{\model}{\textsc{PatchNet}\xspace}
\newcommand{\FH}{feature heterogeneity\xspace}

\newcommand{\vpara}[1]{\vspace{0.05in}\noindent\textbf{#1 }}

%% file: 1-abs.tex
\begin{abstract}
In recent years, the rapid development of foundation models and graph pre-training technologies has spurred increasing interest in constructing a universal pre-trained graph model or Graph Foundation Model (GFM). However, a significant challenge is that existing models are unable to address feature heterogeneity in graph data without textual information, which hinders the transferability of graph models across different datasets.
To bridge this gap, we propose the concept of \textit{learnable graph patches}, which we regard as the smallest semantic units of any graph data.
We decompose the graph into learnable graph patches by unfolding the node features and constructing corresponding patch structures separately.
We then design \model, a framework that mines transferable information from graph data across domains. Specifically, after extracting graph patches, we propose a patch encoder to extract knowledge from each unit and a patch aggregator to learn how the units are combined into a whole.
Due to its domain-agnostic nature, the model can be applied to downstream data across different domains.
Furthermore, we analyze the connection between \model and existing graph models, as well as the transferability of the node embeddings it generates.
Empirically, our method not only achieves the capability to use multi-domain graphs for pre-training, but also shows enhanced performance across various downstream datasets and tasks. Moreover, we observe consistent improvement in downstream performance as the volume of pre-training data increases.
\end{abstract}

%% file: 2-intro.tex
\section{Introduction} \label{sec:intro}

Graph, as a prevalent data structure, is ubiquitous in a wide range of applications, such as social network analysis \cite{yang2019mining}, bio-informatics \cite{yi2022micer}, finance \cite{yang2019loan}, etc.
These graphs form a vast knowledge base, encompassing rich and comprehensive information across various fields.
With the advancement of graph neural networks (GNNs) and pre-training techniques, the development of universal graph pre-train models or graph foundation models (GFMs) that absorb knowledge across diverse graphs has gained significant attention~\cite{mao2024graph,liu2023towards}.

\begin{figure}[ht]
    \centering
    \includegraphics[width=0.48\textwidth]{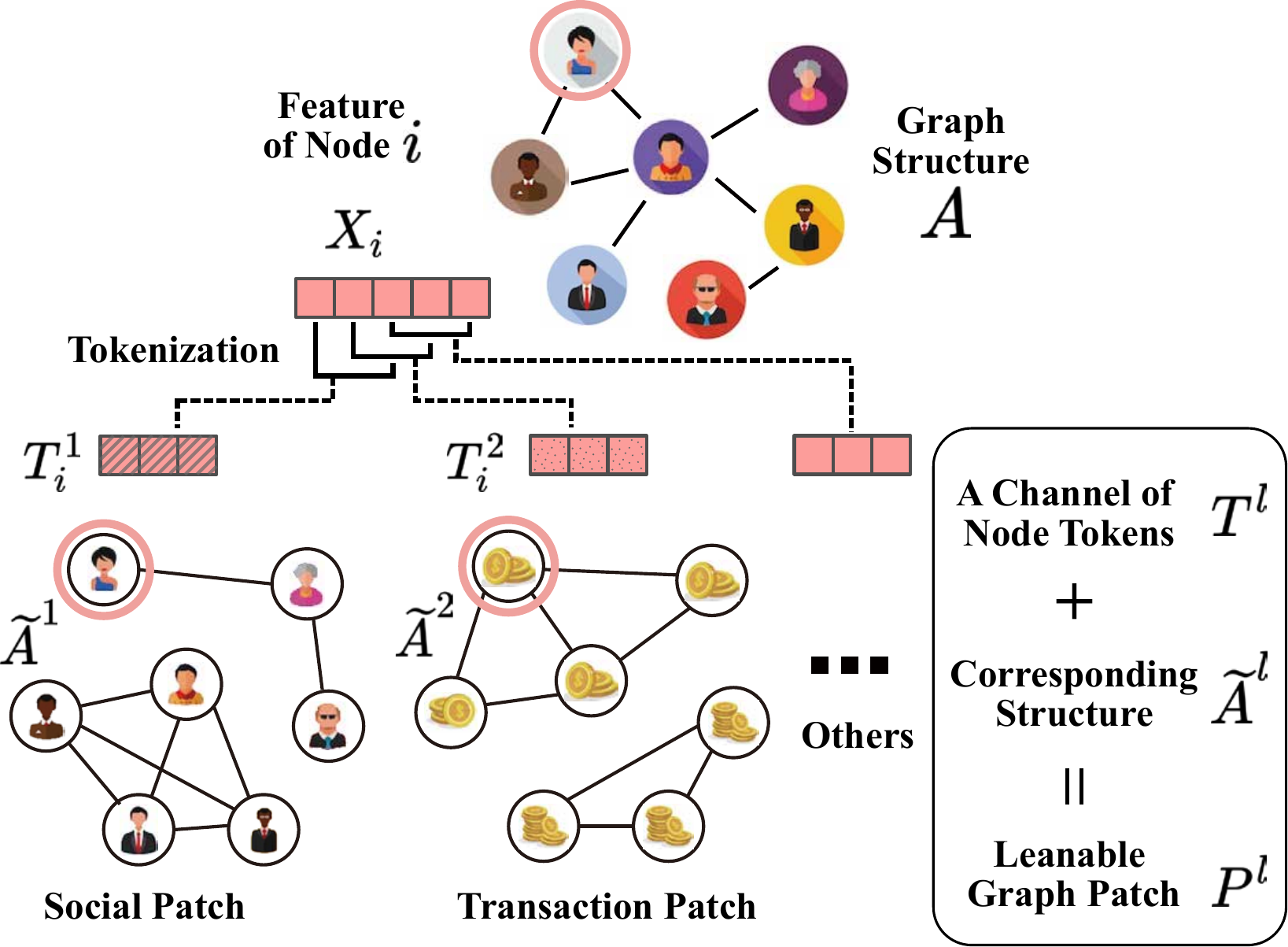}
    \setlength{\belowcaptionskip}{-0.2cm}
    \caption{Decomposing a graph into \textbf{graph patches}. The upper side is an input graph (financial network as example). The lower side shows two of the decomposed graph patches. The first patch represents a social network, utilizing the social attributes of users for its node token, with edges indicating similarities in social aspects. The second patch is a transaction network, featuring nodes characterized by transactional information, where edges reflect transactional similarities.}
    \label{fig:graphpatch}
\end{figure}

However, efforts to develop transferable pre-trained models have faced numerous challenges.
The reason is that graph data exhibits more complex heterogeneity compared to images in computer vision (CV) and sentences in natural language processing (NLP). One reason lies in the structural aspect, graph data exhibits vastly different patterns across various domains. For example, benzene rings are commonly found in molecular graphs, while triangles frequently appear in social networks. Fortunately, this issue has seen considerable breakthroughs in recent research~\cite{qiu2020gcc,zhu2021transfer}.

On the other hand, \FH poses significant challenges and has been less explored due to its complex nature. Specifically, the sources of node information in graph data can be various, and the process of converting raw data into features also varies greatly. This leads to node features in different graph datasets potentially existing in entirely different semantic spaces. For example, node features in molecular graphs might be manually annotated by experts in laboratories without further processing, whereas in financial networks, features could be derived from tabular data using some feature selecting method.

One existing solution involves introducing text feature and using LLMs to map these into a common space~\cite{oneforall}. However, most graph data do not include text as part of the node information, and in some cases, the feature information cannot be provided due to privacy concerns.
Another approach is to use simple singular value decomposition (SVD) to convert feature to the same length~\cite{zhao2024all}, but this only standardizes feature length without aligning the joint distribution of node features and structure across different graphs. Thus, effectively addressing \FH on graphs becomes crucial for building a unified graph pre-training model.

The core challenges in handling \FH include, firstly, the difficulty in extracting meaningful and transferable information from the vastly diverse features across different graph data.
Secondly, how to preserve such transferable information using a scalable model during the pre-training process also presents a significant challenge.
In this paper, we propose that the association of feature structures could be the key to transferability. Hence, we aim to consider both node features and structural information when addressing \FH.
The key idea is to design a \textit{learnable graph patching module} which is adaptable to various kinds of feature and train it to extract the transferable information to boost performance of downstream tasks.

First, we propose that each graph can be seen as a complex object consisting of graph patches.
As the example shown in Fig.~\ref{fig:graphpatch}, the financial network on the upper side can be decomposed into a series of graph patches, including a social graph patch and a transaction graph patch, among others. Each graph patch contains a piece of relatively independent and transferable information from the original graph.
Note that the node tokens $T$ of each graph patch $P$ are unfolded from the original node features, and its graph structure $\widetilde{A}$ is learned from both the original structure and the node tokens.
We propose that such graph patches preserve the basic transferable information between graphs, as other graphs may also be decomposed into similar social and transactional information.

Second, we propose to build a scalable graph model, \model, to encode the captured information in the graph patches.
Specifically, we first extract learnable graph patches from original graph, which are composed of node tokens and corresponding structures.
Since the transferable information is hidden in the graph patches, we design a patch encoder to aggregate graph patch of each channel.
Then we build a patch aggregation module to learn how to combine these transferable information across channels of graph patches.
Our method is the first one that can be pre-trained on data across domains without the requirement for text or other side information, and can be applied to downstream data across domains. Additionally, the high degree of modularity of our model allows for stacking to scale up the model's parameter.
Finally, we evaluate our model on domain transfer settings and self-supervised setting. The results indicate that this is a promising step toward building a GFM.

The contribution of this paper are summarized as follows:
\begin{compactitem}
\item  We propose the concept of learnable graph patches to handle \FH, which enables the preserving of multi-domain knowledge across different graphs.
\item  We propose a graph model, \model, composed of patch encoder and patch aggregator for generalized graph pre-training. We analyze the obtained embedding quality and the correlation between \model and existing models.
\item  We empirically prove the effectiveness of \model in various settings, which demonstrate the superior performance across multiple graph datasets and tasks.
\end{compactitem}

%% file: 3-related.tex
\section{Related Work}
\vpara{Tackling \FH.}
Developing transferable GNNs or GFM has always been a hot topic of research in the graph community.
One of the great challenges is dealing with \FH~\cite{mao2024graph,liu2023towards}.
Existing methods for graph pre-training or transfer learning suffer from \FH due to varying feature origins and semantic spaces.
OFA~\cite{oneforall} manually converts heterogeneous features into textual descriptions using LLMs. However, most of the graphs mainly contain feature attributes without either side information like text or homogeneous origin.
GCOPE~\cite{zhao2024all} employs domain-specific virtual nodes that serve as inter-connectors linking nodes across various domains.
Although these virtual nodes enable connections between domains, they only apply SVD to address \FH without explicitly capturing the transferable information.
However, we propose to use learnable graph patches to capturing transferrable information within \FH.

\vpara{Graph patching or tokenization.}
Due to the complex non-Euclidean nature of graph data, many existing studies have addressed the associated challenges by decomposing graphs into smaller units like patches or tokens to enable more effective modeling and information extraction.
These methods can be categorized into 4 types based on the granularity of the decomposition: node-level, path-level, subgraph-level, and combination-level.
The most common approach is node-level decomposition, a method employed by various graph transformers~\cite{rampavsek2022recipe,dwivedi2020generalization,ying2021transformers,min2022transformer}. The advantage of this level is the natural way of segmentation, although it does not reduce the complexity of the graph structure.
Next is the path level: PathNet~\cite{ijcai2022p310} enhances the discriminability of graph models on heterophily graphs by capturing path patterns, while PathNNs~\cite{michel2023path} theoretically demonstrate that capturing full-range paths can enhance both expressiveness and model performance. The subgraph level includes NAGphormer~\cite{chennagphormer}, which efficiently improves node classification on large graphs by aggregating each layer of neighbors as separate patches, and GPatcher~\cite{zhang2023gpatcher}, which proposes aggregating $p$ neighbors for each node based on topology and node features in heterophily graphs; Graph ViT~\cite{he2023generalization} employs METIS for graph segmentation, validating that ViT-like approaches can also be effective on graphs. The final category is the combination level, to which our method belongs.
We propose that learning such patch is transferable across graphs because these learnable graph patches can learn common unit information across graphs even with \FH.

%% file: 4-model.tex
\section{Method}

\begin{figure*}[ht]
    \centering
    \includegraphics[width=0.9\textwidth]{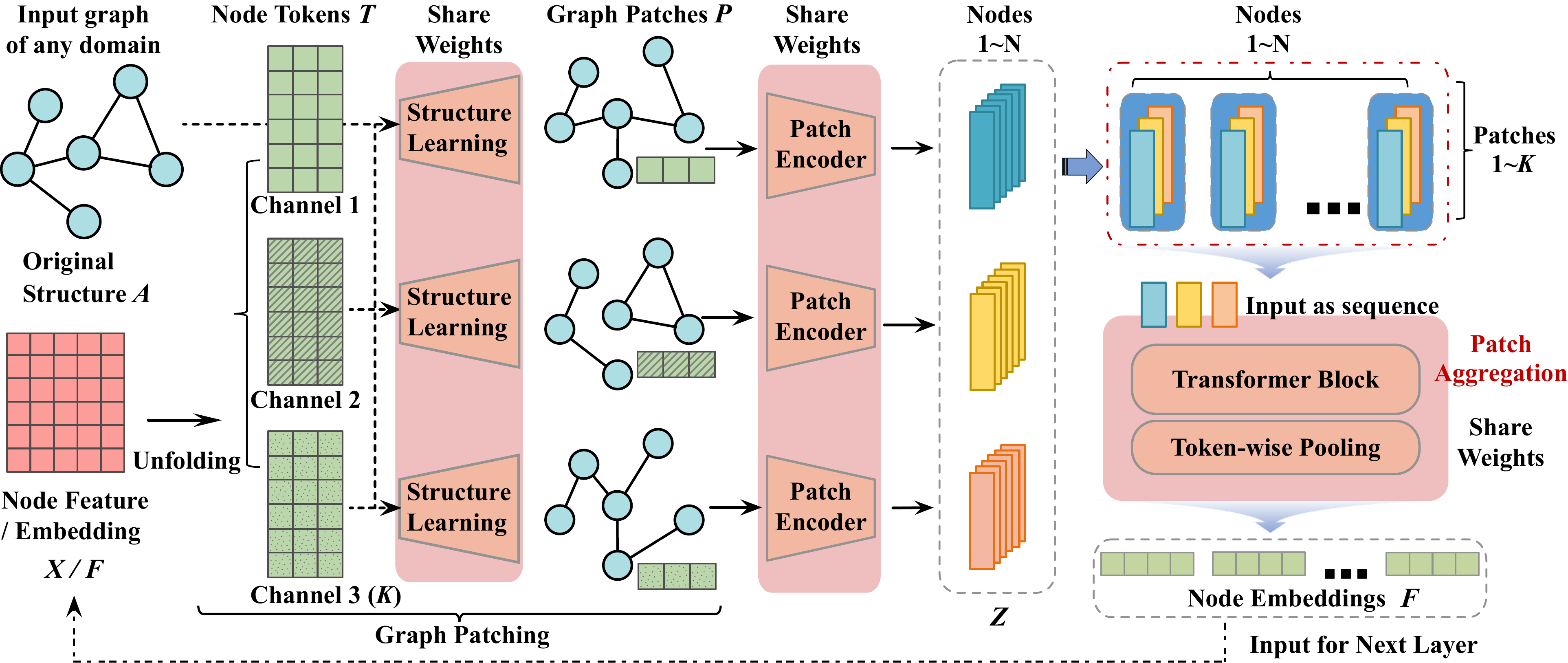}
    \caption{Overall architecture of \model. (a) Building Graph Patches: The input node attributes $X$ are unfolded into multiple node tokens, which are then paired with a graph learner to form patches. (b) Encoding Patches: Each patch is encoded using a shared-parameter encoder, resulting in patch embeddings $Z$ without information passing between patches. (c) Aggregating Patches: Patch embeddings are aggregated within each node to yield the final node embeddings $F$.}
    \label{fig:model}
\end{figure*}

\vpara{Notations.} We denote a graph as $\mathcal{G}=\left(V,A,X\right)$ and $N$ is the number of nodes, $A\in \{0,1\}^{N\times N}$ is the adjacency matrix and $X_\text{ori}\in \mathbb{R}^{N\times D}$ is the node attribute matrix where $D$ is the dimension of attributes that differs across domains. Due to \FH, we normalize the node attributes as $X \in [0,1]^{N \times D}$.

\vpara{Overview.}
The overall architecture of \model is shown in Fig.~\ref{fig:model} and the full algorithm can be found in Appendix.
As discussed in Sec.~\ref{sec:intro}, the goal is to handle \FH between graphs in order to transfer the knowledge from various pre-training datasets into target downstream datasets. Moreover, we propose that the key is to build learnable patches that can capture the inherent feature and structure correlation. After obtaining graph patches, we first perform an encoding within each graph patch to obtain the inner-patch embedding, known as the Patch Encoder. Subsequently, we aggregate across all patches, referred to as Patch Aggregation, to produce the final node embedding $F$, which can be used for node classification or graph classification after pooling. This complete processing represents one layer of \model, meaning that the entire process can be repeated multiple times to capture more extensive and richer patch information.

\subsection{Building Structure-aware Graph Patches} \label{subsec:tokenExtraction}
The construction of graph patches consists of two parts: the formulation of node tokens and patch structure learning.
\begin{definition}[Unfolding Node Tokens] \label{def:tokens}
Given a graph $\mathcal{G}=\left(V,A,X\right)$, the normalized attributes of each node $X_i \in [0,1]^{D}$ can be unfolded into node tokens $T_i \in [0,1]^{{K} \times {M}}$, where $K$ is the number of channels of node tokens and $M$ is the token size.
This can be formalized as follows: $T_i^l = X[i, start_l:end_l], start_l=S \times l, end_l=S \times l+M, l=0,...,K$,  $M$ is the size of each token, step $S$ refers to the stride of the unfolding window, which can be calculated by $S=\lfloor \frac{D-M}{K} \rfloor$.
Here, each token $T^{l}_i \in [0,1]^{{M}}, l \in {1,..., K}$ is a smaller information unit of node.
Each channel of tokens for all nodes is denoted as $T^{l} \in [0,1]^{{N} \times {M}}$.
Hence, the node token matrix of this graph is $T \in [0,1]^{{N} \times {K} \times {M}}$.
\end{definition}

The node token in Fig.~\ref{fig:graphpatch} is represented by the tokenized pink vectors, which are extracted from the original node features.
The core idea is to use the tensor unfold operation to losslessly expand each node's features into tokens.
This unfold operation resembles the first half of a convolution operation, which is akin to spreading out potentially useful information.
The single channel of tokens for all nodes in the graph contains a single perspective of all features.

\vpara{Relation between node tokens.}
We explain the following two relations: The relation between node tokens on the same node $T_i^1, T_i^2$ and the relation between node tokens across different nodes $T_1^l, T_2^l$.
Note that subscripts represent node indices, while superscripts denote token or patch indices.
$T_i^l = X[i, start_l:end_l]$, the neighboring $T_i^1, T_i^2$ share the overlap feature. Since $start_1<start_2<end_1<end_2$  and $end_1-start_2=(M-S)$, the shared part between $T_i^1, T_i^2$ is a subset of token $T_i^1$.
Since the process of splitting tokens from node features does not involve interactions between nodes, the relationship between  $T_1^l, T_2^l$  is similar to that between $X_1,X_2$, meaning that they are relatively independent.

After we formulate tokens from each node with the original structure (dotted edges in Fig.~\ref{fig:model}), we then introduce a structure learning method to derive learnable graph patches based on the original graph structure and node tokens. This part is our key to handling \FH across graphs.

\begin{definition}[Learning Graph Patches] \label{def:patches}
Given a channel of tokens $T^{l}$ and the original graph structure $A$, a graph patch is denoted $P^{l} = \left(T^{l},\widetilde{A}^{l}\right)$, where $l ={1,..., K}$. Here, $\widetilde{A}^{l} \in \{0,1\} ^ {N\times N}$ is learned for each token channel $l$ with a shared graph learner $\Psi$ in parallel. That is, $\widetilde{A}^{l}= \Psi(T^{l}, A)$.
\end{definition}

Thus, a graph can be divided into $K$ learnable graph patches for further modeling.
We propose that the key to tackling \FH is to find/learn the transferable information across different graphs. Here we propose to design a learnable module to extract the graph patches that contain feature-structure correlation information.
In other words, the learned graph patches themselves are not transferable, but the learning mechanism within the parameters of graph learning module is transferable.
Note that the graph learner module are learned simultaneously with patch encoder and aggregator which are introduced in following sections.

There are many ways to construct such a graph learner module. In this paper, we utilize an attention-based approach.
\begin{align}
\label{eq:attentive_learner}
S_{i,j} = f_\phi(W \odot T^{l}_i, W \odot T^{l}_j), \\
\widetilde{A}_{i,j}(T) = \begin{cases} \sigma(S_{i,j}), & j \in \operatorname{top-k} (S_{i,:}) \\ 0, & j \notin \operatorname{top-k} (S_{i,:}) \end{cases},
\end{align}
\noindent where $T^{l}_i$ and $T^{l}_j$ denote two node tokens, $\odot$ is the Hadamard operation, $W$ is a learnable parameter vector, $f_\phi$ denotes the similarity metric such as cosine similarity, $\sigma$ stands for non-linear activation function like relu.
Moreover, we employ residual connections to update the learned $\widetilde{A}$ to accelerate and stabilize the training process:
\begin{equation} \label{eq:residual}
\widetilde{A} = \alpha \widetilde{A}+(1- \alpha ){A}
\end{equation}
\noindent where $\alpha$ stands for the trade-off parameter for how much trainable structure to adopt.

\subsection{Theoretical Analysis of Graph Patches}

\vpara{Fundamental assumption.} The success of transfer learning relies on the pre-training and downstream data having similar distributions as model inputs~\cite{zhuang2020comprehensive}. Compared to the original graph, patches of the same granularity are more likely to exhibit similar distributions, as the extracted tokens have the same dimensions, and the patch structures are learned by a transferable graph learner. This enables the model to transfer learned knowledge effectively when modeling and aggregating patches across different datasets.

\vpara{Physical meaning.} Any graph data is partitioned into learnable graph patches $P^l = (T^l,\widetilde{A}^{l})$ with the same token feature space $T^l \in [0,1]^{{N} \times {M}}$ and the same space of learned structure $\widetilde{A}^{l} \in \{0,1\} ^ {N\times N}$, representing the division into smaller graph patches with the same Cartesian Product space of $T \times \widetilde{A}$. Ultimately, patching reduces the distributional differences between $\Pr(T(G_a), \widetilde{A}(G_a))$  and  $\Pr(T(G_b), \widetilde{A}(G_b))$ of different graphs $G_a, G_b$.

\vpara{Relation among patches.} Note that when generating node tokens $T^1, T^2$ from the original node features, we use an unfolding operation with overlap ($T^1 \cap T^2 \neq \emptyset$). The $\widetilde{A}$ of each patch is learned based on the same $A$. Thus, the information between two adjacent patches (adjacent in the token splitting process) partially overlaps, which means these patches are not entirely independent. On the other hand, since the node tokens obtained through unfolding can be restored via the folding operation ( $\bigcup_{l=1}^{K} T_{i}^l = X_{i}, i=1,...,N$), all the patches can be combined to reconstruct the original graph ( $\bigcup_{l=1}^{K} P^l = G(X, A)$).

\subsection{Encoding and Aggregating Graph Patches}\label{subsec:intra-patch}
In this section, we introduce scalable graph patch encoder and aggregator to extract transferable knowledge in graph patches.

\vpara{Encoding Graph Patches.}
We propose a dual-branch attention mechanism to adaptively encode the patches.
Specifically, given each token $T^l \in [0,1]^{N \times M}$, the original $A$ and learned $\widetilde{A}$, the $l$-th patch are encoded in following two steps:
we first use GNN with shared parameters to respectively encode both the learned patches and the combination of the original graph with node tokens.

\begin{align} \label{eq:dual-branch1}
H^l=\Phi(T^l,A), \quad
\widetilde{H^l}=\Phi(T^l,\widetilde{A}(T^l)),
\end{align}

where $\Phi$ denotes a GNN. This results in embeddings $H^l, \widetilde{H^l}$, both in $\mathbb{R}^{N \times \text{hid}}$.
Then we combine both $H^l$ and $\widetilde{H^l}$ to get the representation of the $l$-th patch of the whole graph.

\begin{align} \label{eq:dual-branch2}
f_{\textbf{MLP}}( \widetilde{H^l} || H^l) = E \in \mathbb{R}^{N \times 2}, \quad
\delta(E)=\beta \in (0,1)^{N \times 2},
\end{align}

\noindent where $f_{\textbf{MLP}}$ is a two-layer MLP and $\delta$ stands for Softmax operation. The Softmax operation is applied along the second dimension to produce normalized importance scores $\beta$.
Finally, we utilizes the scores $\beta$ to combine $H^l, \widetilde{H^l}$.

\begin{align} \label{eq:dual-branch3}
 Z^l= \sum \delta(E),
\end{align}

where the summation is applied along the second dimension.
The output $Z^l \in \mathbb{R}^{N \times \text{hid}} (l=1,...K)$ is encoded patch embedding for $N$ nodes.

\vpara{Aggregating Graph Patches.}
For images or natural languages, after extracting the patches, it is vital to combine with positional embedding before aggregating the patches.
However, the positional embedding of tokens, and even that of each dimension of node attributes, is meaningless for graph data. For example, in a social network where users are nodes, swapping the age and gender attributes does not affect the information of the nodes or the graph itself.
Thus, we directly feed the patch embeddings into the patch aggregation process.
Moreover, since we have node-wisely (along the 1st dimension) aggregated the tokens with graph structure, we propose to patch-wisely (along the 2nd dimension) aggregate the patches for every node.

One straight way is to aggregate all the patches through a MLP-based module. However, since \model is designed to transfer across different domains which cannot guarantee the same number of patches, we propose to employ a module with unlimited capacity of aggregation length, transformer block.
It enables \model to capture contextual information throughout the entire feature.
Given $Z \in \mathbb{R}^{N \times K \times  \text{hid}} $ as input, the
\begin{align}
U &= \text{LayerNorm}(\text{MHA}(Z)), \\
U' &= \text{LayerNorm}(\text{FFN}(U)),  \\
F &= \text{Pooling}(W') \label{eq:pool}
\end{align}
\noindent where $\text{MHA}$ function computes the attention for each of the $h$ heads and concatenates them together, $\text{FF}$ function applies two linear layers with ReLU activation in between, $\text{LayerNorm}$ is used to provide training stability and the \text{Pooling} refers to pooling along the direction of $K$ patches.
Thus, we get the final embeddings $F \in \mathbb{R}^{N \times f} $.
Overall, the time complexity of patch aggregation is $\mathcal{O}({NK^2})$ .

It is noteworthy that, through the aggregation within patches and between patches, only one layer of the proposed framework is completed.
Similar to the concept of the layer in classical GNNs, we can stack multiple layers of \model.
Specifically, when using \model for the first layer, the input features are the original node features. In the second layer of using \model, we perform a new layer of graph patch construction and modeling based on the node embeddings generated in the first layer.

\subsection{Pipeline and Intuition}
In this section, we aim to illustrate the process of using \model during pre-training and fine-tuning, and to explain the intuition behind our model's ability to handle \FH.

\vpara{Detailed Implementation.} Firstly, we shuffle multiple datasets and feed them into \model. We segment all graphs into graph patches and encoding and aggregating graph patches to get node embedding.
Then, we use existing self-supervised tasks to enable \model to learn transferable information across graph datasets.
During downstream fine-tuning, the downstream data must also be segmented according to the patching steps shown in Sec.~\ref{subsec:tokenExtraction}.
The major difference in fine-tuning is that the tasks are switched to downstream tasks. After fine-tuning on the training data, we use the tuned \model to infer on the test set.

Here we give the analysis of time complexity of \model. The pre-processing the tokens takes $\mathcal{O}(NKM)$, where number of patches is $K$ and token size is $M$. The encoding of structure-aware graph patches is divided into two parts: For the structure learning part, we account for complexities involving locality-sensitive kNN sparsification post-processing~\cite{fatemi2021slaps}, where neighbors of each node are selected from a batch of nodes (batch size = $b$, hidden dimension $\approx M$). Thus, the complexity is $\mathcal{O}(NM^2 + NMb)$~\cite{liu2022towards}. The GNN aggregation typically takes $\mathcal{O}(N+E)$.
Lastly, patch Aggregation takes $\mathcal{O}(NK^2)$.
Hence, the overall complexity is: $\mathcal{O}(N(KM+M^2+Mb+K^2))$.

\vpara{Intuition.}
Here we intuitively explain why the proposed graph patch learning is transferable to tackle \FH.
The success of transfer learning relies on the pre-training and downstream data having similar distributions as model inputs~\cite{zhuang2020comprehensive}. Compared to the original graph, patches of the same granularity are more likely to exhibit similar distributions, as the extracted tokens have the same dimensions and the patch structures are learned by a transferable graph learner. This enables the model to transfer learned knowledge effectively when modeling and aggregating patches across different datasets.
Note that Fig.~\ref{fig:graphpatch} is merely illustrative, assuming that in a financial network, some features represent social information while some represent transaction information. In reality, the token features in the data may not always correspond to clearly defined or linguistically describable aspects.
Thus, we use the the learnable parameters in structure learning module to automate the learning of the patch construction.

Since we have learned the transferable information across datasets during patch extraction, we further preserve these information by encoding and aggregating the patches.
Then, we use encoder to acquire knowledge within graph patches with the same information granularity. Subsequent transformer-based patch aggregation learns how different graphs combine these graph patches.
In summary, the information transferred by \model is encapsulated within the extracted graph patches. The richer the variety of graph patches during pre-training, the more likely it is that similar graph patches will be encountered in downstream data, thus enhancing downstream performance.

\subsection{Model Analysis}\label{subsec:theory}

\vpara{Connection to others.} Compared to the existing graph pre-training backbone, where only transferable information could be learned from a complete graph, dividing the graph into patches allows for learning both fine-grained inner-patch patterns and correlations between coarse-grained inter-patches information. In this section, we analyze the correlation between \model, existing MPNNs and other graph transformers.

\begin{figure}
    \centering
    \includegraphics[width=.43\textwidth]{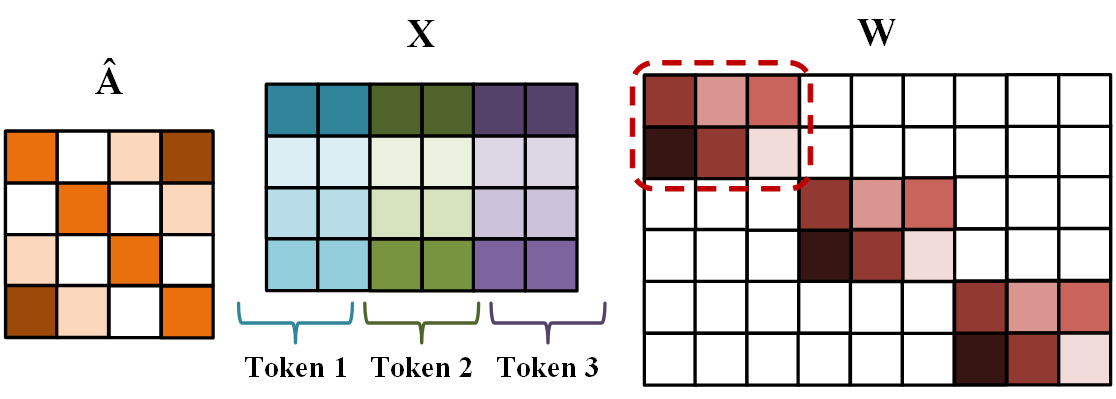}
    \caption{The decomposed operation of \model from the perspective of MPNN.}
    \label{fig:analysis}
    \vspace{-0.35cm}
\end{figure}

Theoretically, \model can be seen as a decoupled version of existing MPNNs. Formally, each layer of an MPNN can be formulated as the combination of ``propagation'' and ``transformation'':
\begin{equation}
    \mathbf{H} = \sigma\big(\mathbf{\hat{A}}\mathbf{X}\mathbf{W}\big), \quad \mathbf{\hat{A}} = \widetilde{\mathbf{D}}^{-\frac{1}{2}}\widetilde{\mathbf{A}}\widetilde{\mathbf{D}}^{-\frac{1}{2}},
\end{equation}
\noindent where $W$ is the weight matrix of the ``transformation'' process.
As shown in Fig.~\ref{fig:analysis}, our patch encoder decouples the aggregation process between patches, which was originally achieved through aggregation with graph structure $\hat{A}$ in MPNNs, and implements it through the transformer block in Section~\ref{subsec:intra-patch}. Thus, our patch encoder can be seen as setting the parameters of the white part in the $W$ matrix to zeros.
These blank parameters are replaced by the powerful transformer block, which not only retains the learning ability but also reduces the noise impact of the graph structure on the misaligned patches.
In addition, since our structure-aware module shares weights for $K$ patches, the parameters near the diagonal of our $W$ matrix are duplicated and identical, which is shown to be parameter-efficient.
From another perspective, $W$ in Fig.~\ref{fig:analysis} is essentially a block diagonal matrix. Such matrices are often used in scenarios aimed at enhancing efficiency while still maintaining the accuracy of the algorithm~\cite{dao2022monarch}.

Generally, \model reduces information exchange between patches compared to MPNN in the patch encoding stage and earlier steps, which we compensate for during patch aggregation. In other words, our method retains the flexibility of parameter transformations on input features, similar to MPNN, while significantly surpassing MPNN in terms of generalizability.

\begin{figure}
    \centering
    \includegraphics[width=0.5\textwidth]{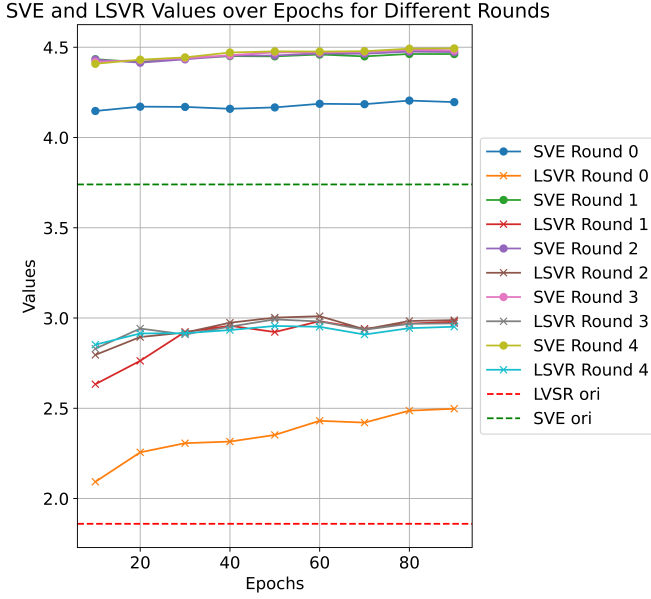}
    \setlength{\abovecaptionskip}{-0.05cm}
    \caption{The SVE and LSVR of the embedding generated by \model.}
    \label{fig:obs}
    \vspace{-0.2cm}
\end{figure}

\vpara{Quality of embedding.} To further analyze \model, we propose to qualify the generated node embedding from the perspective of singular value.
The singular value spectrum of the embedding space, which is widely considered to be related to the generalization performance \cite{oymak2019generalization, chen19itrans, xue2022investigating}.
More specifically, we perform singular value decomposition (SVD) on the node embedding $\mathbf{F} \in \mathbb{R}^{M \times D}$ by \model:
$\mathbf{F} =\mathbf{U} \boldsymbol{\Sigma} \mathbf{V}^{\top}$.\footnote{ $\mathbf{U}$ and $\mathbf{V}$ denotes the left and right singular vector matrices, respectively, and $\mathbf{\Sigma}$ denoting the diagonal singular value matrix $\{\sigma_1, \ldots, \sigma_D\}$.}.

\begin{definition}[Singular Value Entropy]
Singular value entropy (SVE) is characterized as the entropy associated with the normalized singular values. It serves as a quantifier for the distribution's flatness among singular values.
\begin{equation}
     \mathrm{SVE} = -\sum_{i=1}^D \frac{\sigma_i}{\sum^{D}_{j=1} \sigma_j} \log \frac{ \sigma_i}{\sum^{D}_{j=1} \sigma_j}
\end{equation}
\end{definition}
Higher SVE values suggest an enhanced capture of data structure within the feature space, attributed to either the learning of more distinct features or the memorization of noise, thereby expanding its dimensional span.

\begin{definition}[Largest Singular Value Ratio]
 The largest singular value ratio (LSVR) is determined by taking the logarithm of the quotient obtained from dividing the largest singular value, denoted as $\sigma_1$, by the aggregate of all singular values:
\begin{equation}
    \mathrm{LSVR} = -\log \frac{\sigma_1}{\sum^{D}_{i=1} \sigma_i}.
\end{equation}
LSVR quantifies the disparities in data encapsulated by the singular vector associated with the largest singular value, $\sigma_1$, indicative of the model's transferability~\cite{chen19itrans}.
\end{definition}

We plot the SVE and LSVR for the embedding generated by \model in Fig.~\ref{fig:obs}. The input data here is PCQM4Mv2~\cite{hu2021ogblsc}. Specifically, the dashed lines represent the average values of SVE and LSVR obtained from the original data's node features, while the different colored solid lines represent the average values of SVE and LSVR for the node representations obtained by our model after different numbers of rounds of training. The solid lines with dots represent the values of SVE, and the solid lines with crosses represent the values of LSVR. The results indicate that during the forward propagation process of \model, both SVE and LSVR for the node representations are continuously increasing, demonstrating that our model is constantly improving the transferability and distinguishability of the representations.

%% file: 5-exp.tex
\begin{figure*}[ht!]
\begin{minipage}{0.63\textwidth}
    \centering
\captionof{table}{Cross domain pre-training and fine-tuning performance in terms of mean and std. deviation of ROC-AUC (for Sider, HIV, Bace) and F1 (for Flickr and DBLP). Improvement (IMP) and P-value are used to measure the gap between using \model with all pre-training data and without any pre-training.}
\label{tab:cross}
\resizebox{1.0\textwidth}{!}{
\begin{tabular}{c|c|c c c | c c }
\toprule
Model & \diagbox{\small Pre}{\small Down}  & Sider & HIV & Bace & Flickr & DBLP  \\
\midrule
GIN & {No pre-train}
  &  52.14$\pm$0.56 & 56.58$\pm$2.57 & 55.84$\pm$3.16   & 47.25$\pm$3.54 & 74.86$\pm$2.26  \\
\model & {No pre-train}
   &  51.88$\pm$0.59 & 57.72$\pm$1.31 & 57.35$\pm$2.77   & 46.29$\pm$3.86 & 75.05$\pm$2.52  \\ \hline
\multirow{3}{*}{\model} & {ZINC}
  & \underline{53.06$\pm$0.47} & \underline{58.74$\pm$1.72} & \underline{59.74$\pm$0.54}  & 45.30$\pm$2.98 & 76.44$\pm$2.32  \\
 & {Arxiv}
  & 52.77$\pm$0.58 & 57.01$\pm$1.80  & 56.00$\pm$2.66 & \underline{50.64$\pm$2.69} & \underline{79.01$\pm$2.84}   \\
 & {ZINC+Arxiv}
 & \textbf{53.71$\pm$0.48} & \textbf{60.23$\pm$1.15} & \textbf{59.75$\pm$1.02}   & \textbf{50.81$\pm$1.9}4 & \textbf{79.12$\pm$ 1.93} \\  \hline
 \multicolumn{2}{c|}{IMP (\%)} & 1.29 &  2.51  &  2.40 &  4.52  &  4.07  \\
 \multicolumn{2}{c|}{ P-value } & 0.012 &  0.008  &  0.010 &  0.003 & 0.003 \\ \bottomrule
\end{tabular}
}
\end{minipage}
\hfill
\begin{minipage}{0.36\textwidth}
\centering
\includegraphics[width=\textwidth]{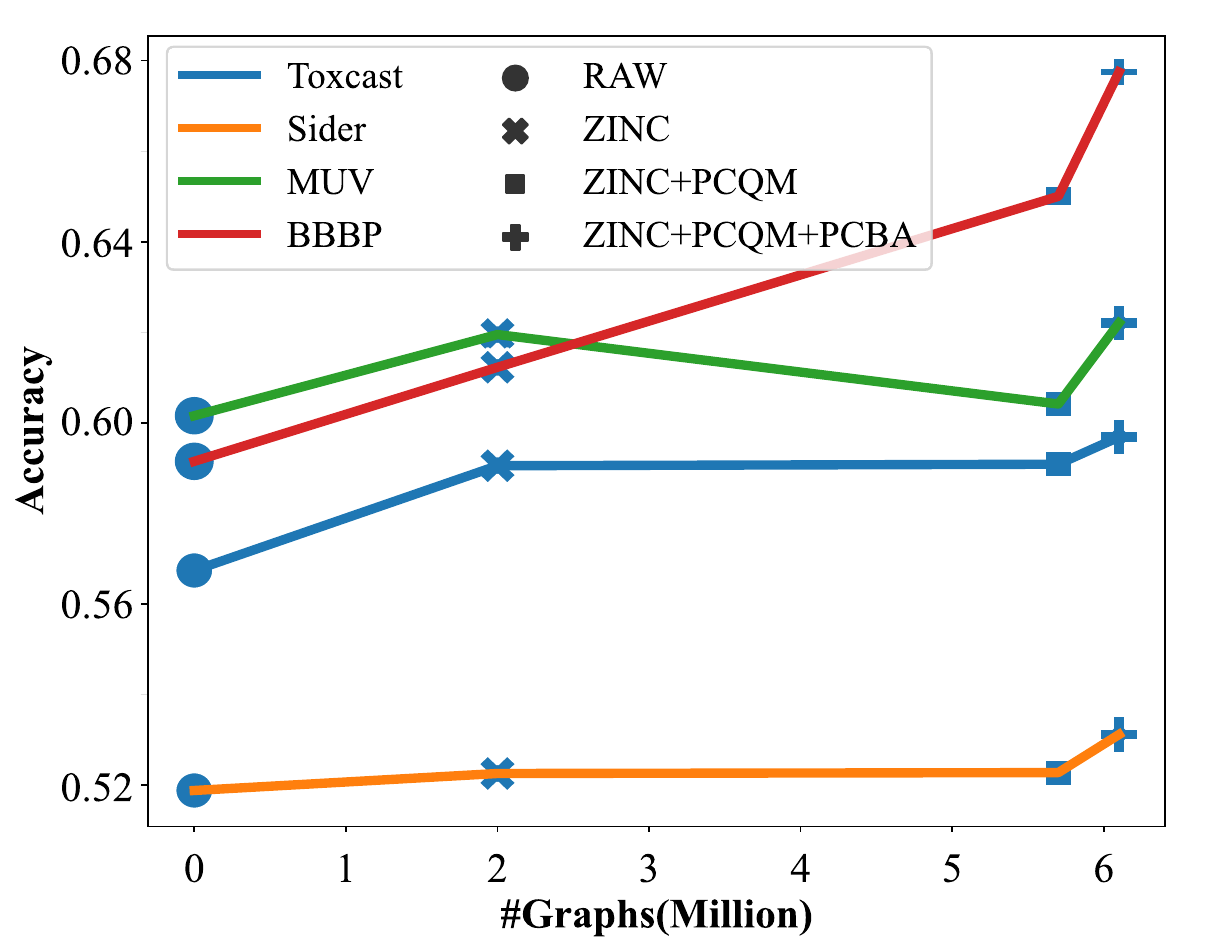}
\setlength{\abovecaptionskip}{-0.04cm}
\captionof{figure}{Performance on scaling of the size fo pre-training datasets.}
\label{fig:analysis1}
\end{minipage}
\end{figure*}

\section{Experiments}
In this section, we answer the following four questions through experiments to validate the effectiveness of \model\footnote[2]{Our codes are available at \url{https://github.com/zjunet/PatchNet}.}:
\begin{itemize}
    \item \textbf{RQ1.} Can \model handle feature heterogeneity on different downstream datasets through cross-domain pre-training?
    \item \textbf{RQ2.} Does the performance of \model improve with the increase of the scale of pre-training datasets?
    \item \textbf{RQ3.} Can \model outperform other pre-trained backbones in both graph and node classification tasks?
    \item \textbf{RQ4.} How sensitive is \model to the size of node tokens?
\end{itemize}
Note that each set of experiments is repeated five times. Detailed hyper-parameters can be found in the Appendix.

\subsection{Cross Domain Transfer Learning}
To address \textbf{RQ1}, we set up experiments (Tab.~\ref{tab:cross}) with cross-domain pre-training and fine-tuning.
\textit{Some graph data's features are derived from text and are not the target of this paper, as they can either be generated using the same text encoder to obtain homogeneous features, or use models like OFA~\cite{oneforall} with LLMs.}
To the best of our knowledge, no other models have yet been pre-trained across multiple domains and then applied to different tasks in various downstream domains, so we only use \model to evaluate.

During pre-training, we selected molecular dataset ZINC~\cite{sterling2015zinc} and paper citation dataset Ogbn-Arxiv~\cite{hu2020ogb}. These datasets not only exhibit feature heterogeneity but also have different forms: Arxiv is a single large graph, whereas ZINC consists of many smaller molecular graphs. Thus, we employ neighbor sampling for Arxiv and graph sampling for ZINC. We load these two datasets simultaneously to pre-train one \model using existing two pre-text tasks, Attribute Masking (AM) and Context prediction (CP)~\cite{hu2019strategies}. We empirically find out that using these multi-task trick~\cite{lin2019pareto} to combine AM and CP performs the best. Note that since ZINC has 2M samples (graphs) while Arxiv only has 0.17M samples (subgraph stems from nodes), we randomly choose 0.17M samples out of ZINC in each epoch. The reason is to keep balance between the pre-training information. In the fine-tuning phase, we employed two different types of tasks across datasets: graph and node classification; the graph classification involved Sider, HIV, and Bace from \cite{wu2018moleculenet}, while the node classification included datasets Flickr~\cite{zeng2019graphsaint} and DBLP~\cite{8970680}. Notably, the downstream data also originated from different domains: Sider, HIV, and Bace are molecular datasets, while Flickr and DBLP are social network and citation graph, respectively. Not only is there feature heterogeneity between the pre-train and downstream datasets, but they also belong to different tasks. During fine-tuning, we conduct fine-tuning and testing according to the data splits specific to each dataset. In Tab.~\ref{tab:cross}, each row represents a combination of pre-training data, and each column represents a different downstream dataset. The first two rows represent end-to-end learning without any transfer.

\vpara{Results.} From Tab.~\ref{tab:cross}, it is evident that our model handles \FH effectively, both between pre-training datasets and between pre-training and downstream datasets. Large IMP values demonstrate that our method has achieved performance improvements through pre-training and transfer learning, while small p-values ensure that these improvements are statistically significant.  Moreover, although \model may sometimes perform comparably to basic models like GIN~\cite{xu2018how} (used without pre-training) in an end-to-end manner, it consistently shows improvement after pre-training. Furthermore, the underlined scores representing the second-best results generally indicate that pre-training with the same task data tends to yield relatively better results. Additionally, we find that even when the two pre-training datasets come from completely different backgrounds, combined pre-training still leads to improvements in downstream tasks. Even when there is significant variation among downstream datasets, this type of cross-domain transfer pre-training still achieves good results in downstream tasks. This is also the outcome we hope to see in future graph foundation models, positioning \model as a step towards the realization of a GFM.

\begin{table*}[htbp]
\resizebox{\textwidth}{!}{
\begin{tabular}{l|cccccccc|c}
\toprule
 &
  Tox21 &
  Toxcast &
  Sider &
  ClinTox &
  MUV &
  HIV &
  BBBP &
  Bace &
  Rank \\ \hline
GIN (w/o pre)      & 67.90$\pm$1.48 & 58.39$\pm$0.96 & 52.14$\pm$0.56 & 56.43$\pm$4.23 & 58.53$\pm$2.52 & 56.58$\pm$2.57 & 58.57$\pm$7.72 & 55.84$\pm$3.16 & 12.5
     \\ \hline

ContextPred~\cite{hu2019strategies} &
 66.45$\pm$1.75  &
 58.16$\pm$0.85  &
 51.53$\pm$0.22  &
 55.83$\pm$1.07  &
  59.49$\pm$2.66 &
  56.58$\pm$1.31 &
  63.57$\pm$1.16 &
 57.92$\pm$1.07  & 11.3
   \\
AttrMask~\cite{hu2019strategies} &
  67.17$\pm$1.31 &
  59.33$\pm$1.37 &
 52.21$\pm$1.12  &
 56.69$\pm$1.78  &
  58.58$\pm$2.18 &
  57.34$\pm$0.98 &
 63.65$\pm$2.18  &
   57.27$\pm$1.94 & 9.3
   \\
AM+CP~\cite{lin2019pareto} &
  67.62$\pm$1.84 &
  58.19$\pm$0.68 &
  52.44$\pm$0.29 &
  57.17$\pm$0.96 &
  59.06$\pm$2.63 &
  56.53$\pm$1.43 &
  63.79$\pm$1.60 &
  57.96$\pm$3.64 & 9.3
   \\
GPT-GNN~\cite{hu2020gpt} &
67.98$\pm$1.75   &
58.39$\pm$1.51   &
52.97$\pm$0.91   &
57.07$\pm$1.73   &
58.56$\pm$1.54   &
56.68$\pm$1.03   &
65.06$\pm$3.05   &
56.25$\pm$2.05   & 8.5
   \\
GraphCL~\cite{you2020graph} &
  68.22$\pm$1.61 &
 59.09$\pm$1.18  &
  52.67$\pm$0.09 &
 56.99$\pm$1.63 &
58.73$\pm$1.85  &
 56.82$\pm$1.64  &
   64.68$\pm$1.44 &
  56.92$\pm$1.26 & 7.9
   \\
   \hline
GraphMVP~\cite{liu2021pre} &
  68.01$\pm$0.93 &
  55.43$\pm$0.44 &
  52.24$\pm$0.57 &
  55.54$\pm$2.12 &
  57.36$\pm$2.69 &
  56.88$\pm$1.75 &
  65.41$\pm$0.39 &
  57.77$\pm$0.35 &10.3
   \\
3D InfoMax~\cite{stark20223d}&
 67.05$\pm$1.26  &
 58.22$\pm$0.58 &
  52.58$\pm$0.35  &
  54.56$\pm$2.55 &
 59.85$\pm$2.36  &
 56.65$\pm$1.67  &
  67.64$\pm$1.33 &
  58.66$\pm$1.40 & 9.0
   \\
Mole-BERT~\cite{xia2023mole}    & \textbf{70.07$\pm$0.69} & \textbf{59.72$\pm$0.15} & 52.58$\pm$0.35 & 55.52$\pm$3.09 & 61.05$\pm$1.35 & \underline{57.79$\pm$1.79} & \underline{68.44$\pm$2.90} & \textbf{60.07$\pm$1.71} & \underline{4.0} \\ \hline
\model (w/o pre) & 67.77$\pm$1.80  & 56.74$\pm$0.59 & 51.88$\pm$0.59 & 55.78$\pm$1.15 & 60.16$\pm$1.17 & 57.72$\pm$1.31 & 59.15$\pm$2.85 & 57.35$\pm$2.77 & 10.4 \\
\model (AM) & 64.84$\pm$1.41 & 56.95$\pm$0.74 & 52.29$\pm$0.39 & 57.26$\pm$2.81 & 60.65$\pm$2.50 & 54.27$\pm$1.39 & 60.32$\pm$4.27 & 57.20$\pm$1.94 & 11.1 \\
\model (CP) & 66.22$\pm$1.83 & 59.56$\pm$0.27 & \underline{53.06$\pm$0.52} & \underline{58.60$\pm$1.35} & \underline{61.79$\pm$1.47} & 54.11$\pm$1.46 & 55.01$\pm$6.11 & 57.55$\pm$1.34 &  7.9 \\
\model (AM+CP) &
  \underline{68.56$\pm$1.34} &
  \underline{59.64$\pm$0.71} &
  \textbf{53.06$\pm$0.47} &
  \textbf{59.50$\pm$1.79} &
  \textbf{62.19$\pm$1.46} &
  \textbf{58.74$\pm$1.72} &
  \textbf{68.55$\pm$3.75} &
  \underline{59.74$\pm$0.54} & 1.4 \\
\bottomrule
\end{tabular}%
}
\caption{The comparison on graph classification task. "w/o pre" means without pre-training (Full version see Appendix).}
\label{tab:graph_class_main}
\vspace{-0.2cm}
\end{table*}

\begin{table*}[htbp]
\resizebox{0.8\textwidth}{!}{%
\begin{tabular}{l|cccccc|c}
\toprule
 &   Tox21 &   Toxcast &   Sider &   ClinTox &    BBBP &   Bace &   Rank \\ \hline
GCN~\cite{welling2016semi}      & 55.25 $\pm$ 1.68 & 52.14 $\pm$ 0.44 & 51.94 $\pm$ 0.26 & 50.96 $\pm$ 3.73 & 65.27 $\pm$ 1,76  & 53.62 $\pm$ 0.92 & 5.3 \\
GIN~\cite{xu2018how}      &   60.17 $\pm$ 1.84 &   54.19 $\pm$ 0.68 &   51.44 $\pm$ 0.29 &   53.59 $\pm$ 0.96 &    68.10 $\pm$1.60 &   {50.02 $\pm$ 3.64} &  5.0 \\
GAT~\cite{2017Graph}      & 61.62 $\pm$ 2.37 & 53.86 $\pm$ 0.46 & 51.83 $\pm$ 0.28 & 58.41 $\pm$ 2.08 &  68.14 $\pm$ 2.11  & 54.01 $\pm$ 1.73 & 3.8 \\ \hline
Graphormer~\cite{ying2021do}       & 63.04 $\pm$ 1.47 & 57.14 $\pm$ 0.70 &  52.54$\pm$ 0.19 & 55.61 $\pm$ 1.60 &  66.24 $\pm$ 1.62 & 57.56 $\pm$ 1.85 & 3.2 \\
GraphGPS~\cite{rampavsek2022recipe}       & 65.77 $\pm$ 1.78 & 56.13 $\pm$ 0.78 &  52.80$\pm$ 0.21 & 58.75 $\pm$ 1.13 &  \textbf{68.73 $\pm$ 2.17} & 52.11 $\pm$ 1.76   & \underline{2.5} \\ \hline
\model &   \textbf{68.56 $\pm$ 1.34} &   \textbf{59.64 $\pm$ 0.71} &   \textbf{53.06 $\pm$ 0.47} &   \textbf{59.50 $\pm$ 1.79} &   \underline{68.55 $\pm$ 3.75} &   \textbf{59.74 $\pm$ 0.54} & 1.2 \\
\bottomrule
\end{tabular}%
}
\caption{The comparison of backbones on graph classification task.}
\vspace{-0.2cm}
\label{tab:backbone}
\end{table*}

\begin{table}[htbp]
	\centering
{
	\begin{tabular}
{ c | c  c  }
\toprule
    & Coauthor-CS & Coauthor-physics \\ 		\midrule
GCN~\cite{welling2016semi}             & 82.15$\pm$1.93     & 87.25$\pm$0.72 \\
GAT~\cite{2017Graph}             & 82.72$\pm$1.96     & 88.41$\pm$0.52  \\
\midrule
DGI~\cite{dgi}             & 83.09$\pm$2.02     & 88.34$\pm$0.75\\
GRACE~\cite{grace}           & 83.43$\pm$2.96     & 88.69$\pm$0.87 \\
GraphMAE~\cite{graphmae}        & 84.33$\pm$2.75    & 90.13$\pm$0.57  \\
GraphMAE2~\cite{graphmae2}       &  84.67$\pm$2.43    & 91.80$\pm$0.64  \\
\midrule
\model &\textbf{87.69$\pm$1.28} &\textbf{92.87$\pm$0.27}\\
\bottomrule
\end{tabular}
}
\caption{Results of node classification evaluation.}
\label{tab:cs+phy}
\vspace{-0.2cm}
\end{table}

\begin{table*}[ht]
\resizebox{0.8\textwidth}{!}{%
\begin{tabular}{l|cccccccc}
\toprule
             & Tox21     & Toxcast   & Sider     & ClinTox   & MUV       & HIV       & BBBP      & Bace \\
\midrule
\model &  \textbf{68.56$\pm$1.34} &
  \textbf{59.64$\pm$0.71} &
  \textbf{53.06$\pm$0.47} &
  \textbf{59.50$\pm$1.79} &
  \textbf{62.19$\pm$1.46} &
  \textbf{58.74$\pm$1.72} &
  \textbf{68.55$\pm$3.75} &
  \textbf{59.74$\pm$0.54} \\ \hline
Inter-Mean & 60.87$\pm$1.50 & 53.41$\pm$0.86 & 52.11$\pm$0.90 & 51.46$\pm$1.26 & 59.35$\pm$1.09 & 55.91$\pm$2.20 & 50.58$\pm$3.08 & 52.01$\pm$0.54   \\
Inter-Sum  & 60.65$\pm$2.17 & 51.48$\pm$0.80 & 52.80$\pm$0.43 & 58.79$\pm$1.30 & 58.05$\pm$1.92 & 52.34$\pm$3.07 & 52.02$\pm$7.82 & 53.89$\pm$0.93   \\ \hline
Inner-GCN  & 63.78$\pm$2.10 & 55.93$\pm$0.53 & 51.62$\pm$0.47 & 50.13$\pm$1.66 & 58.74$\pm$1.17 & 53.82$\pm$1.70 & 54.04$\pm$5.18 & 53.18$\pm$2.02   \\
Inner-GAT   & 59.14$\pm$3.76 & 56.07$\pm$1.34 & 52.54$\pm$0.60 & 57.98$\pm$1.23 & 59.79$\pm$1.27 & 58.21$\pm$3.21 & 54.19$\pm$5.83 & 54.84$\pm$1.58   \\ \hline
Non-overlap  & 60.62$\pm$2.36 & 54.68$\pm$0.92 & 52.89$\pm$0.41 & 58.31$\pm$1.54 & 61.47$\pm$1.86 & 56.60$\pm$2.39 & 53.71$\pm$4.57 & 55.54$\pm$1.59 \\
\bottomrule
\end{tabular}%
}
\caption{Ablation on our backbone}
\label{tab:ablation}
\vspace{-0.3cm}
\end{table*}

\subsection{Data Scaling Perspective}
To address \textbf{RQ2}, we control the domain variable, increasing only the scale of the pre-training dataset within a single domain.
We use different combinations of pre-training datasets to increase the volume of pre-training data.
Fig.~\ref{fig:analysis1} uses pre-training data that includes ZINC, PCQM~\cite{hu2021ogblsc}, and PCBA~\cite{hu2020ogb}. By combining these for pre-training, we obtain three models, plus one model without pre-training, resulting in four data points of each line. The downstream data includes Toxcast, Sider, MUV and BBBP from ~\cite{wu2018moleculenet}, corresponding to four colors of lines.

\vpara{Results.} From Fig.~\ref{fig:analysis1}, we find that \model is capable of multi-dataset pre-training and efficiently handling larger amounts of pre-training data. Additionally, certain pre-train datasets contribute more significantly to improvements in specific downstream datasets compared to others. For example, PCBA has a greater impact on enhancing MUV compared to its impact on Toxcast and Sider. This may be attributed to the fact that both PCBA and MUV are classified within the Biophysics category, whereas Sider is categorized under Physiology data~\cite{wu2018moleculenet}. Moreover, \model performs better when pre-trained with more datasets, which reveals that our model has great potential when trained with enormous data.

\subsection{Pre-training and Fine-tuning}
To address \textbf{RQ3}, we conduct experiments on two types of downstream tasks: graph classification and node classification.

\vpara{Graph classification evaluation.}
In this section, we conduct two parts of experiments: comparing with different pre-text tasks (Tab.~\ref{tab:graph_class_main}) comparing with different backbones (Tab.~\ref{tab:backbone}).
We choose ZINC as our pre-training datasets in both experiments.
Note that the size of node feature is $D=300$, which is different from the publicly available feature size $D=2$. It's well known that most of the researches involving molecular graphs all expand two-dimensional original feature to 300 dimensions of learnable features~\cite{xia2022systematic}.
Thus, we replace the learnable feature with non-trainable features by using group VQ-VAE~\cite{oord2017neural} which is also used by Mole-bert~\cite{xia2023mole} which are frozen throughout the training process.

For patch extraction, we use token size $M = 32$ and step size $K = 20$.
For downstream evaluation, we adopt the widely-used 8 binary classification datasets from MoleculeNet~\cite{wu2018moleculenet}.
Due to the difficulty of obtaining sufficient labels in practical applications, we adopt a 1:1:8 train-validation-test label split .
Note that we employ scaffold splitting~\cite{ramsundar2019deep} to split molecules based on their structures, which emulates real-world scenarios.

On the one hand, we fix the backbone of baselines as their original GIN.
Tab.~\ref{tab:graph_class_main} is divided into 4 main blocks by row. The first row is using GIN as backbone and without pre-training.
The second block consists of current popular pre-training strategies. The third block contains those strategies designed for molecular graphs. In the last block we adopt 3 combination of pre-text tasks: attribute masking, context prediction, and both.
From Tab.~\ref{tab:graph_class_main}, we can see that:
Compared to all baselines, \model achieves competitive or better performance under the same experimental protocols. Since we only use simple pre-training strategies and its combination, it is apparent that our backbone plays the most significant role.
Moreover, we find that single pre-training strategies may lead to negative transfer on both GIN and \model. But after applying multi-task pre-training strategy, \model opens up a significant gap with GIN, which means our model does better on combining pre-training strategies.
When testing \model's raw capability, we are surprised to find that \model outperform some GIN series models in certain datasets even without pre-training, which indicates that even under the condition of such extreme label ratio and more model parameters, \model still achieves better performance, reflecting that \model is adaptive and robust.
Due to space limitation, we show the full version in Appendix.

On the other hand, we compare with different backbones using the same pre-training strategy as the combination of AM and CP.
Tab.~\ref{tab:backbone} is also divided into 3 blocks, universal backbones, molecular specialized backbones and our backbone from top to bottom.
From Tab.~\ref{tab:backbone}, we can see that:
As is known to all, the effect of transfer between pre-train and downstream datasets is an important measurement for backbones. \model provides a well-performed transfer compared to traditional backbones such as GCN, GAT and GIN. This is because patch extraction process significantly increases the generalizability of our model, and the transformer module can avoid the old problems such as over-smoothing, allowing our model to effectively use a greater amount of parameters than normal GNNs.
Moreover, our method performs better than molecule specialized backbones such as Graphormer. That's because we combine the strengths of both GNN aggregation and transformer.
Since our backbone can obtain more detailed semantic information in original features by learning from tokens, it is much more efficient and effective than other backbones.

\vpara{Node classification evaluation.}
Here we use two datasets, Coauthor-CS~\cite{shchur2018pitfalls} and Coauthor-Physics~\cite{shchur2018pitfalls} for both pre-training and downstream evaluation due to their rich node features.
Tab.~\ref{tab:cs+phy} contains 3 blocks: classical end-to-end methods, well-known self-supervised methods and \model.
As for dataset split, we follow ParetoGNN~\cite{ju2023multi} with 1:1:8 for training/validation/test.
We report the average ROC-AUC with the corresponding standard deviation.
Tab.~\ref{tab:cs+phy} shows that our backbone is more powerful and capable of extracting more information from the limited downstream data. Since \model is trained on both datasets, our method has achieved cross-dataset transferability, which enables us to pre-train a model using more large-scale datasets to get a more powerful model.

\begin{figure}[htbp]
    \centering
    \includegraphics[width=0.33\textwidth]{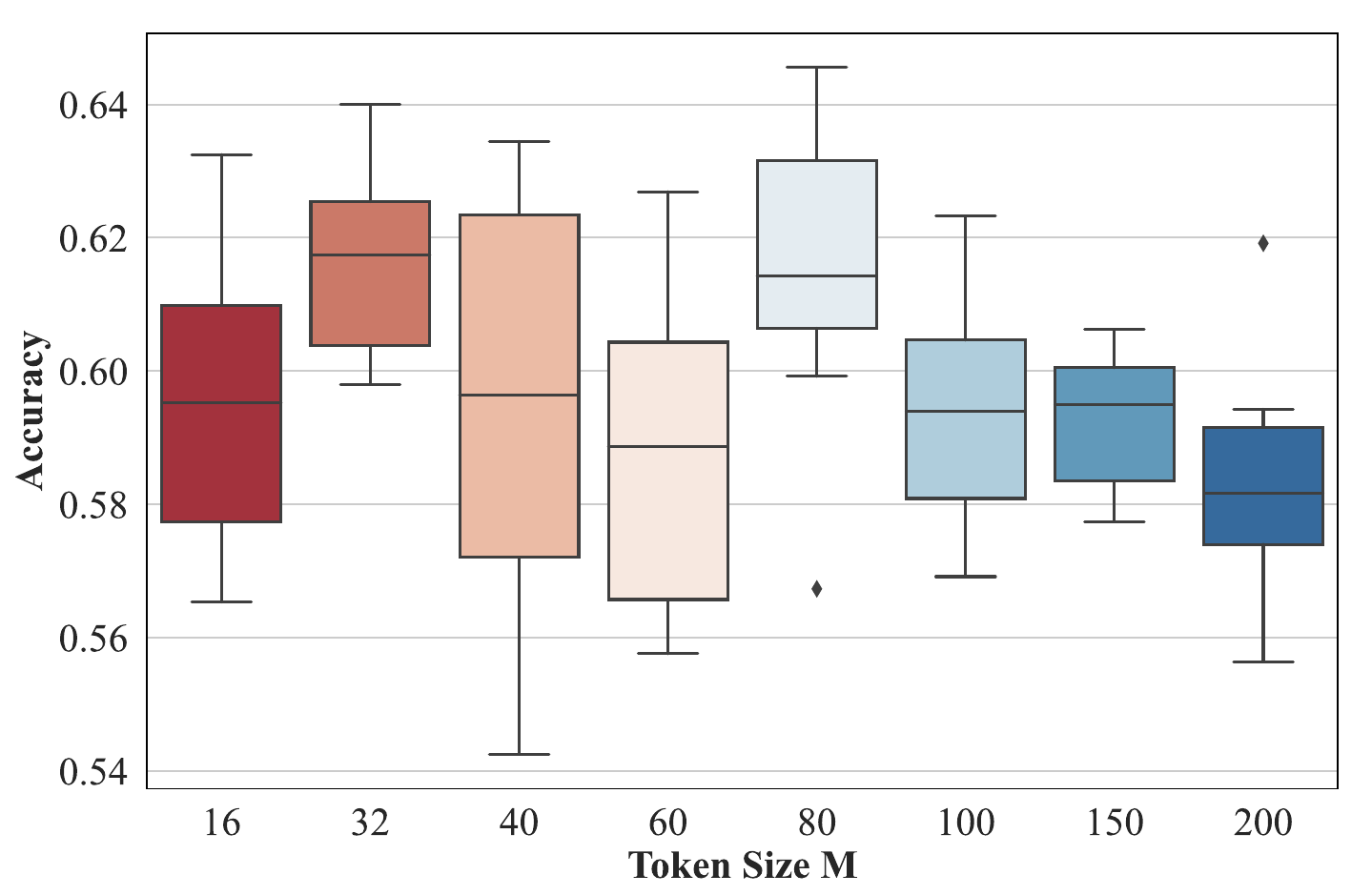}
    \caption{ Performance of varying token size $M$.}
\vspace{-0.2cm}
    \label{fig:analysis2}
\end{figure}

\subsection{Ablation and Sensitivity Study}
To address \textbf{RQ4}, we conduct two experiments: an ablation study of \model' s sub-modules and a sensitivity analysis of the most critical hyperparameter, the token size $M$.

As shown in Tab.~\ref{tab:backbone}, the backbone with transformer performs significantly better than that with simple pooling, since the rich information among different patches cannot be simply aggregated by pooling.
And the inner-patch aggregation with GIN does a better work than others. This is due to the fact that GIN has high expressiveness,  which is also the reason why GIN has become the preferred choice in many molecular graph-based research studies.
What's more, we test the behavior of our model when there is no overlapping information between patches. Since splitting original feature of any dataset is quite challenging, we allow overlapping tokens to form graph patches. And as our expectation, the backbone with token overlapping performs better. But to our surprise, some of the results are better than the setting without pre-training, which means our model could potentially learn to complete missing features.
From Fig.~\ref{fig:analysis2}, we find that having a large number of token size isn't a good idea. This is because if the number of token channels is relatively low, the learning process for aggregating information between tokens will be partially lost. In extreme scenarios where there's only one token, this process will be lost entirely, which means our backbone will devolve into a to normal GNN model. Moreover, we can find out that even when token size is small, satisfactory results can still be achieved. It indicates that GNN's ability to learn structure is not abandoned even when the backbone focuses more on transformer's token aggregation.

%% file: 6-conclusion.tex
\section{Conclusion}

In summary, we highlight the challenges in addressing the limitations of current graph pre-training models for the incapability of handling cross-domain transfer. We identify the main challenge as \FH. Then, by introducing the learnable graph patches as a basic semantic unit for graph data, we propose \model, a framework capable of mining transferable information across different graph domains.
Our empirical analyses show \model's capability to generate distinguishable and transferable node representations, advancing pre-training on multi-domain non-textual graphs and showing continuous improvement on various downstream datasets and tasks.

\vpara{Limitation and future work.} Our method has an inherent limitation in using a fixed size for unfolding node features into node tokens, determined by a hyperparameter based on empirical experience. Future work will explore automatic learning of optimal node token sizes to improve generalization.

\section{Acknowledgment}
This work is supported by National Natural Science Foundation of China (No. 62176233, No. 62322606, No. 62441605).

%% file: 7-appendix.tex
\appendix

\section{Details of Experiments}

\subsection{Datasets} \label{app:data}
The datasets used are shown in Tab.~\ref{tab:graph_classification_hyper}.
The scale of the graphs used in pre-training and downstream tasks also highlights the high scalability of \model.

\begin{table}[htbp]
\renewcommand\arraystretch{1.2}
\centering
 \resizebox{0.4\textwidth}{!}
 {
    \setlength\tabcolsep{2pt}
    \begin{threeparttable}
    \begin{tabular}{cccccc}
     \toprule
        & \multicolumn{1}{c}{Type}  &  \multicolumn{1}{c}{Name}  &  \multicolumn{1}{c}{$N$}  &  \multicolumn{1}{c}{$E$}     \\
     \midrule
     \multirow{4}*{\setlength\tabcolsep{1pt}\rotatebox{90}{\textbf{Pre-training}}} & {\textcolor[rgb]{0.9,0.2,0.3294}{\textbf{Graph level}}}  &  ZINC~\cite{sterling2015zinc} &  53,254,058  &  115,472,818   \\
     & ~   &  PCQM4Mv2~\cite{hu2021ogblsc} & 52,970,652 & 109,093,626   \\
     & ~   &  Ogbg-molpcba~\cite{hu2020ogb} & 11,373,137  & 24,618,372   \\
     & {\textcolor[rgb]{0.765,0.4078,0.145}{\textbf{Node level}}}  & Ogbn-arxiv~\cite{hu2020ogb} & 169,343 & 1,166,243   \\
     & ~ &Flickr~\cite{zeng2019graphsaint} & 	105,938 & 2,316,948   \\
     & ~ &DBLP~\cite{8970680} & 28,702 & 68,335  \\
     \midrule
     \multirow{13}*{\rotatebox{90}{\textbf{Downstream}}}
     & {\textcolor[rgb]{0.9,0.2,0.3294}{\textbf{Graph level}}} &  Tox21~\cite{wu2018moleculenet}  & 145,459  & 302,190  \\
      & ~ & Toxcast~\cite{wu2018moleculenet} & 161,088 & 330,356  \\
       & ~ & Sider~\cite{wu2018moleculenet} & 48,006 & 100,912  \\
        & ~ & ClinTox~\cite{wu2018moleculenet} & 38,637 & 82,372  \\
         & ~ & MUV~\cite{wu2018moleculenet} & 2,255,846 & 4,892,252   \\
         & ~ & HIV~\cite{wu2018moleculenet} & 1,049,163 & 2,259,376   \\
         & ~ & BBBP~\cite{wu2018moleculenet} & 49,068 & 105842  \\
         & ~ & Bace~\cite{wu2018moleculenet} & 51,577 & 111,536   \\
     & {\textcolor[rgb]{0.765,0.4078,0.145}{\textbf{Node level}}}
    &Flickr~\cite{zeng2019graphsaint} & 	105,938 & 2,316,948   \\
     & ~ &DBLP~\cite{8970680} & 28,702 & 68,335  \\
     & ~ & Coauthor-CS~\cite{shchur2018pitfalls} &  18,333 &  81,894  \\
     & ~ & Coauthor-Physics~\cite{shchur2018pitfalls} &  34,493 &  247,962  \\
    \bottomrule
    \end{tabular}
    \end{threeparttable}
 }
\caption{The statistics of all datasets.}
\label{tab:dataset}
\vspace{-0.2cm}
\end{table}

\subsection{Experiment Settings}

\subsection{Hyperparemeter Strategy} \label{app:hyper}

Overall, our proposed framework is implemented via PyTorch. As for software versions, we use Python 3.7.0, PyTorch 1.13.1, OGB 1.3.6, and CUDA 11.3.
Moreover, the range of hyperparameters are listed in Table~\ref{tab:graph_classification_hyper}.

\begin{table}[htbp]
\centering
{
\begin{tabular}{l|c}
\toprule
        Hyperparameter        & Range               \\
\midrule
     $K$        & \{15 $\to$ 25\}       \\
     $M$        & 64 \\
     number of GIN layers & 3 \\
     number of Attention heads & 3 \\
     Learning Rate  & \{1e-3 $\to$ 3e-3\}  \\
     Weight decay   & \{0 $\to$ 1e-6\}  \\
     GIN dropout rate   & 0.2   \\
     Attention dropout rate & \{0.3 $\to$ 0.7\} \\
     Batch size     & \{256, 512, 1024\}  \\
    \midrule
     Optimizer  & Adam  \\
     Epoch      & 100 \\
     GPU        & GeForce RTX 4090 \\

\bottomrule
\end{tabular}
}
\caption{Hyper-parameter of graph classification task. }
\label{tab:graph_classification_hyper}
\vspace{-0.2cm}
\end{table}

\begin{table*}[htbp]
\resizebox{0.86\textwidth}{!}{
\begin{tabular}{l|cccccccc|c}
\toprule
 &
  Tox21 &
  Toxcast &
  Sider &
  ClinTox &
  MUV &
  HIV &
  BBBP &
  Bace &
  Rank \\ \hline
GIN(w/o pre)      & 67.90$\pm$1.48 & 58.39$\pm$0.96 & 52.14$\pm$0.56 & 56.43$\pm$4.23 & 58.53$\pm$2.52 & 56.58$\pm$2.57 & 58.57$\pm$7.72 & 55.84$\pm$3.16 & 12.5
     \\ \hline

AD-GCL~\cite{suresh2021adversarial} &
64.81$\pm$1.22   &
55.55$\pm$0.79   &
51.13$\pm$0.17  &
53.46$\pm$2.20   &
59.41$\pm$2.42   &
56.01$\pm$1.02   &
59.26$\pm$1.86   &
52.27$\pm$2.08   & 15.3
   \\
ContextPred~\cite{hu2019strategies} &
 66.45$\pm$1.75  &
 58.16$\pm$0.85  &
 51.53$\pm$0.22  &
 55.83$\pm$1.07  &
  59.49$\pm$2.66 &
  56.58$\pm$1.31 &
  63.57$\pm$1.16 &
 57.92$\pm$1.07  & 11.3
   \\
AttrMask~\cite{hu2019strategies} &
  67.17$\pm$1.31 &
  59.33$\pm$1.37 &
 52.21$\pm$1.12  &
 56.69$\pm$1.78  &
  58.58$\pm$2.18 &
  57.34$\pm$0.98 &
 63.65$\pm$2.18  &
   57.27$\pm$1.94 & 9.3
   \\
AM+CP~\cite{lin2019pareto} &
  67.62$\pm$1.84 &
  58.19$\pm$0.68 &
  52.44$\pm$0.29 &
  57.17$\pm$0.96 &
  59.06$\pm$2.63 &
  56.53$\pm$1.43 &
  63.79$\pm$1.60 &
  57.96$\pm$3.64 & 9.3
   \\
SimGRACE~\cite{xia2022simgrace} &
 67.49$\pm$1.37  &
 58.79$\pm$0.34  &
 52.73$\pm$0.08  &
 56.54$\pm$2.34  &
  59.98$\pm$2.45 &
   57.65$\pm$1.96 &
63.27$\pm$2.30  &
  56.03$\pm$1.84 & 8.8
   \\
GraphLoG~\cite{xu2021self} &
64.09$\pm$1.47   &
58.88$\pm$0.59   &
52.74$\pm$0.55   &
57.38$\pm$1.76   &
60.39$\pm$1.99   &
57.04$\pm$1.07   &
62.49$\pm$2.80   &
56.14$\pm$1.88   & 8.6
   \\
GPT-GNN~\cite{hu2020gpt} &
67.98$\pm$1.75   &
58.39$\pm$1.51   &
52.97$\pm$0.91   &
57.07$\pm$1.73   &
58.56$\pm$1.54   &
56.68$\pm$1.03   &
65.06$\pm$3.05   &
56.25$\pm$2.05   & 8.5
   \\
GraphCL~\cite{you2020graph} &
  68.22$\pm$1.61 &
 59.09$\pm$1.18  &
  52.67$\pm$0.09 &
 56.99$\pm$1.63 &
58.73$\pm$1.85  &
 56.82$\pm$1.64  &
   64.68$\pm$1.44 &
  56.92$\pm$1.26 & 7.9
   \\
JOAO~\cite{you2021graph} &
  68.41$\pm$1.59  &
 58.92$\pm$0.42  &
 52.45$\pm$0.14  &
 57.78$\pm$1.36  &
 60.73$\pm$2.06  &
56.88$\pm$1.45  &
 62.99$\pm$2.29  &
 57.35$\pm$1.46  & 7.0
   \\ \hline

GraphMVP~\cite{liu2021pre} &
  68.01$\pm$0.93 &
  55.43$\pm$0.44 &
  52.24$\pm$0.57 &
  55.54$\pm$2.12 &
  57.36$\pm$2.69 &
  56.88$\pm$1.75 &
  65.41$\pm$0.39 &
  57.77$\pm$0.35 &10.3
   \\
3D InfoMax~\cite{stark20223d}&
 67.05$\pm$1.26  &
 58.22$\pm$0.58 &
  52.58$\pm$0.35  &
  54.56$\pm$2.55 &
 59.85$\pm$2.36  &
 56.65$\pm$1.67  &
  67.64$\pm$1.33 &
  58.66$\pm$1.40 & 9.0
   \\
Mole-BERT~\cite{xia2023mole}    & \textbf{70.07$\pm$0.69} & \textbf{59.72$\pm$0.15} & 52.58$\pm$0.35 & 55.52$\pm$3.09 & 61.05$\pm$1.35 & \underline{57.79$\pm$1.79} & \underline{68.44$\pm$2.90} & \textbf{60.07$\pm$1.71} & \underline{4.0} \\ \hline
\model(w/o pre) & 67.77$\pm$1.80  & 56.74$\pm$0.59 & 51.88$\pm$0.59 & 55.78$\pm$1.15 & 60.16$\pm$1.17 & 57.72$\pm$1.31 & 59.15$\pm$2.85 & 57.35$\pm$2.77 & 10.4 \\
\model(AM) & 64.84$\pm$1.41 & 56.95$\pm$0.74 & 52.29$\pm$0.39 & 57.26$\pm$2.81 & 60.65$\pm$2.50 & 54.27$\pm$1.39 & 60.32$\pm$4.27 & 57.20$\pm$1.94 & 11.1 \\
\model(CP) & 66.22$\pm$1.83 & 59.56$\pm$0.27 & \underline{53.06$\pm$0.52} & \underline{58.60$\pm$1.35} & \underline{61.79$\pm$1.47} & 54.11$\pm$1.46 & 55.01$\pm$6.11 & 57.55$\pm$1.34 &  7.9 \\
\model(AM+CP) &
  \underline{68.56$\pm$1.34} &
  \underline{59.64$\pm$0.71} &
  \textbf{53.06$\pm$0.47} &
  \textbf{59.50$\pm$1.79} &
  \textbf{62.19$\pm$1.46} &
  \textbf{58.74$\pm$1.72} &
  \textbf{68.55$\pm$3.75} &
  \underline{59.74$\pm$0.54} & 1.4 \\
\bottomrule
\end{tabular}%
}
\caption{The comparison of overall performance on graph classification task.}
\label{tab:graph_class_main1}
\vspace{-0.2cm}
\end{table*}

\subsection{Additional Experiments}
Here we present the complete graph evaluation results in Tab.~\ref{tab:graph_class_main1}, comparing our model against a broader range of self-supervised tasks. It is evident that our model indeed outperforms existing self-supervised methods.
Please refer to the main text for the simplified Tab.~\ref{tab:graph_class_main}.
As is shown in the table, our model's performance, combined with simple pre-training tasks, surpasses other existing approaches on datasets within a single domain. This result demonstrates that our model not only exhibits strong generalization capabilities but also effectively learns domain-specific knowledge.

\section{Algorithm} \label{app:algo}
The algorithm of Fig.~\ref{fig:model} is shown in Algo.~\ref{alg:model_new}. The computation of \model includes three parts: node tokenization, patch construction, encoding and aggregation of graph patches.
Hence, the overall complexity is: $\mathcal{O}(N(KM+M^2+Mb+K^2))$.

\begin{algorithm}[h!]
    \caption{Pseudo code for the forward process of the model}
    \label{alg:model_new}
    \begin{itemize}
        \item[] The Patch Encoder $\Phi$, the multi-head self-attention encoder $W$ and a feed forward network using ReLU $F$. An Attention Mechanism $Att$.
        \item[] The graph structure and node feature after patch extraction is ${g, x}$. If we regarded each channel separately, we will get ${g, x_i}, i=0..K$ for different tokens $i$.
        \item[] /* Model training starts */
        \item[] Obtain their node embeddings $z_0$ by [Some Method].
        \item[] \textbf{for} each token $j$ \textbf{do}
        \begin{itemize}
            \item[] Learn the new graph structure $g'$.
            \item[] \textbf{for} each node $u$, node $v$ \textbf{do}
            \begin{itemize}
                \item[] Calculate similarity $S_{uv}^j = SIM(W \cdot x_{uj}, W \cdot x_{vj})$.
            \end{itemize}
            \item[] Apply non-linear transformation to $S^j$.
            \item[] Normalize $S^j$.
            \item[] Update graph structure $g' \gets Topk(g, S^j)$.
            \item[] Combine node embeddings using attention mechanism: $z_{0,j} \gets Att(\Phi(g, x_j), \Phi(g', x_j))$.
        \end{itemize}
        \item[] \textbf{end for}
        \item[] Aggregate through patches by $W$ and pass through $F$: $z_1 \gets W(z_0)$.
        \item[] Combine embeddings: $z_2 \gets F(z_0 + z_1)$.
        \item[] Merge all channels by mean-pooling: $z_3 \gets P(z_0 + z_1 + z_2)$.
    \end{itemize}
\end{algorithm}

\section{More Related Work}
We provide additional information here on related work concerning pre-training and fine-tuning GNNs.
The concept of graph foundation model (GFM) is comprehensively established~\cite{liu2023towards,mao2024graph} and envisioned to be adept across various tasks and datasets.
Yet, there currently does not exist a GFM that fully meets the criteria.
However, building a cross-domain and cross-task graph model has always been a hot topic.
One of the pathways to build graph foundation models is to design graph pre-training framework.

Pre-training graph models has achieved significant success, with various self-supervised pre-training methods proposed for both node-level and graph-level tasks.
W2PGNN~\cite{cao2023pre} explores how to construct pretraining graph datasets to maximize downstream performance gains.
Following the generative language model GPT \cite{radford2019language}, GPT-GNN \cite{hu2020gpt} factorizes graph generation into Attribute Generation and Edge Generation.
GraphCL \cite{you2020graph} maximizes agreement between two representations of the same node by injecting random perturbations, and \cite{hu2019strategies} use subgraphs to develop several self-supervised learning strategies combining node-level and graph-level pre-training information. Graph Contrastive Coding (GCC) \cite{qiu2020gcc} captures the universal network topological properties through subgraph instance discrimination as pre-training task.
Recent self-supervised learning on graph data~\cite{Fang2024ExploringCO} explores the relationships between different tasks and designs models to achieve the most balanced embeddings for labels.
However, a majority of these works still utilize a plain GNN, such as the 5-layer Graph Isomorphism Network(GIN)~\cite{xu2018how,xia2022survey}, and therefore cannot be reused when encountering different downstream tasks without corresponding data.

Besides, some other works utilize the uniqueness of input data in designing architectures.
For example, MolCLR \cite{wang2022molecular} uses molecule SMILES to implement graph augmentation for contrastive learning. GraphMVP \cite{liu2022pretraining} pretrains by leveraging the consistency between 3D geometry and 2D topology.
Since these models utilize more semantic information, they are even more domain-specific.
Some of above methods only transfer structural information, neglecting the node attributes that contain valuable information.

After pretraining, fine-tuning is often required to adapt the pre-trained graph model to downstream data and tasks. GPF~\cite{Fang2023UniversalPT} is the first to introduce efficient prompt tuning into graph model. G-tuning~\cite{sun2024fine} and Bridge-Tune~\cite{huang2024measuring} bridge the gaps between pre-training and downstream from the perspectives of data and tasks, respectively, during the tuning stage.

Recently, attempts have been made to adapt LLMs to tasks associated with graph analysis. Despite their proficiency in natural language processing, directly converting graph data for LLMs processing has not been entirely effective, leading to less than ideal outcomes, as demonstrated in research on both textual~\cite{chen2023exploring} and non-textual graphs~\cite{wang2023can}. Nevertheless, LLMs still encounter challenges when processing graph data~\cite{saparov2022language,dziri2023faith}.